\documentclass[letterpaper, 10 pt, conference]{ieeeconf}  % Comment this line out if you need a4paper

\IEEEoverridecommandlockouts                              % This command is only needed if 
\title{\LARGE \bf
Driving in Corner Case: A Real-World Adversarial Closed-Loop Evaluation Platform for End-to-End Autonomous Driving
}

\usepackage{marvosym}

\author{Jiaheng Geng$^{1}$, Jiatong Du$^{1}$, Xinyu Zhang$^{1}$, Ye Li$^{1}$, Panqu Wang$^{2}$, Yanjun Huang$^{1}$\textsuperscript{\Letter}% <-this % stops a space
\thanks{$^{1}$School of Automotive Studies, Tongji University, Shanghai, China.}
\thanks{$^{2}$ZERON, Jiangsu, China.}
\thanks{\textsuperscript{\Letter} Corresponding author}
}
\usepackage{cite}
\usepackage{amsmath,amssymb,amsfonts}
\usepackage{algorithmic}
\usepackage{graphicx}
\usepackage{textcomp}
\usepackage[table,xcdraw]{xcolor}

\usepackage{booktabs}
\usepackage{threeparttable}
\usepackage{caption}
\usepackage{multirow}
\usepackage{array}
\usepackage{bm}
\usepackage{makecell}
\usepackage{epstopdf}
\usepackage{soul}
\usepackage{color}
\usepackage{float}
\usepackage{subfigure}
\usepackage{footnote}
\usepackage{url}
\usepackage{enumerate}
\usepackage{mathrsfs}

\usepackage{stfloats}
\usepackage{amsmath}

\usepackage{hyperref}

\begin{document}

\maketitle
\thispagestyle{empty}
\pagestyle{empty}

%%%%%%%%%%%%%%%%%%%%%%%%%%%%%%%%%%%%%%%%%%%%%%%%%%%%%%%%%%%%%%%%%%%%%%%%%%%%%%%%
\begin{abstract}

Safety-critical corner cases, difficult to collect in the real world, are crucial for evaluating end-to-end autonomous driving. Adversarial interaction is an effective method to generate such safety-critical corner cases. While existing adversarial evaluation methods are built for models operating in simplified simulation environments, adversarial evaluation for real-world end-to-end autonomous driving has been little explored. To address this challenge, we propose a closed-loop evaluation platform for end-to-end autonomous driving, which enables adversarial interactions in real-world scenes. In our platform, the real-world image generator cooperates with an adversarial traffic flow to evaluate various end-to-end models trained on real-world data. The generator, based on flow matching, efficiently and stably generates real-world images according to the traffic environment information. The surrounding vehicle trajectories are selected from a multimodal model trained on large-scale real-world scenarios, and through adversarial score-based filtering, safety-critical corner cases that end-to-end autonomous driving struggles to handle are effectively identified. Experimental results demonstrate that the platform can generate realistic driving images efficiently. Through evaluating the end-to-end models such as UniAD and VAD, we demonstrate that under adversarial traffic flow, our platform evaluates the performance degradation of the tested model in corner cases. This result indicates that this platform can effectively uncover the model's potential issues, which will facilitate the safety and robustness of end-to-end autonomous driving.

\end{abstract}

%%%%%%%%%%%%%%%%%%%%%%%%%%%%%%%%%%%%%%%%%%%%%%%%%%%%%%%%%%%%%%%%%%%%%%%%%%%%%%%%
\section{INTRODUCTION}

End-to-end autonomous driving has attracted significant attention in both academia and industry. By directly mapping inputs such as images to driving actions within a unified framework \cite{chib2023recent, chen2024end}, this paradigm simplifies the system while enhancing generalization \cite{hu2023planning, jiang2023vad, wang2021end}. End-to-end autonomous driving has become one of the most active research directions in autonomous driving.

For autonomous driving, safety-critical corner cases can identify the boundary of model capability and are important for model evaluation. However, corner cases are difficult to collect in the real world \cite{hanselmann2022king,rempe2022generating}. Many studies have shown that developing adversarial surrounding vehicles can efficiently generate safety-critical corner cases \cite{wang2021advsim, ding2021multimodal, chen2021adversarial}. Existing autonomous driving simulators can control traffic flows to generate adversarial scenarios\cite{dosovitskiy2017carla, li2022metadrive, krajzewicz2012recent}. But the corner cases created by these simulators suffer from a significant sim-to-real gap in visual realism: the model trained on real-world datasets cannot be evaluated. Therefore, it is imperative to establish an adversarial closed-loop evaluation platform for real-world scenes.

To address the sim-to-real gap in visual realism, many researchers employ driving image generation methods. Some studies employ world models to predict future real-world scenes \cite{hu2023gaia, wang2024driving, wang2024drivedreamer}, where historical latent information is propagated to estimate future states and decode them into future images. However, such methods often suffer from weak controllability over the generated images, while adversarial scenarios require strong controllability over surrounding vehicles, which is clearly contradictory. Diffusion-based approaches for driving dataset generation enable conditional controllability\cite{swerdlow2024street, yang2023bevcontrol, wen2024panacea, gao2023magicdrive}. Challenger\cite{xu2025challenger} creates conditions that induce adversarial scenarios, thereby generating adversarial nuScenes-like datasets. Nevertheless, it still lacks closed-loop evaluation, as the ego has no interactivity. DriveArena\cite{yang2024drivearena} integrates SUMO\cite{krajzewicz2012recent} with an image generation model to achieve closed-loop simulation, but the surrounding vehicles are not adversarial. As a result, the discovery of corner cases remains inefficient, and the testing routes are cluttered with redundant and stable traffic conditions. In summary, current methods lack adversarial interactions and cannot create effective corner cases for end-to-end autonomous driving closed-loop evaluation. 

Many researchers generate corner cases by manipulating surrounding vehicles to interact adversarially with the ego. A common approach is to employ reinforcement learning (RL) to enable surrounding vehicles to learn policies that challenge the ego \cite{kuutti2020training, koren2021finding, wang2024efficient, ding2020learning}. Wang et al.\cite{wang2024efficient} leverage RL combined with dynamic and static scene exploration to rapidly search for adversarial safety-critical cases. L2C\cite{ding2020learning} integrates generative models with RL to efficiently generate such scenarios. Moreover, methods like KING\cite{hanselmann2022king} and STRIVE\cite{rempe2022generating} adopt trajectory optimization through cost-function backpropagation, endowing basic trajectory prediction algorithms with adversarial capabilities. However, RL-based approaches are often constrained by their training scenes, limiting their application, and gradient backpropagation is computationally expensive and inefficient \cite{zhang2023cat}. These impose a heavy burden on efficient closed-loop evaluation with real-world image generation. Recently, multimodal trajectory prediction methods have offered new directions for adversarial policy \cite{zhang2023cat, zhang2025co}. In this work, by selecting multimodal trajectories predicted by the model trained on large-scale real-world traffic data and ranking them with adversarial scores, surrounding vehicles can efficiently identify challenging interactions.

% In this work, by filtering trajectories based on their adversarial scores, the surrounding vehicle can efficiently get the adversarial policy.

% By filtering trajectories based on their output scores, these methods can achieve both realism and adversarial effectiveness.

% For closed-loop simulation in real-world scenarios, efficiency is of paramount importance, and reducing the number of denoising steps is a key approach to accelerating image generation for diffusion models \cite{liu2022flow, zhou2024fast, dockhorn2022genie}. Diffusion model generation process still requires tens to hundreds of denoising steps \cite{lu2022dpm}. Reducing the number of steps often degrades the quality of generation \cite{liu2022pseudo}, thereby undermining the realism of the simulation. Flow matching \cite{lipman2022flow} offers a promising solution to this trade-off. By reformulating the stochastic differential equation (SDE) of the diffusion process into a deterministic ordinary differential equation (ODE) \cite{liu2022flow}, it enables high-quality image generation under a few denoising steps, thus significantly improving both the efficiency and the realism of closed-loop simulation in real-world scenarios.

For closed-loop simulation in real-world scenarios, efficiency is of paramount importance. Reducing the number of denoising steps is a key approach to accelerating image generation for diffusion models \cite{liu2022flow, zhou2024fast, dockhorn2022genie}, but the generation process still requires tens of denoising steps \cite{lu2022dpm}. Reducing steps often degrades the quality of generation \cite{liu2022pseudo}, thereby undermining the realism of the simulation. Flow matching \cite{lipman2022flow} offers a promising solution to this trade-off. By reformulating the stochastic differential equation (SDE) of the diffusion process into a deterministic ordinary differential equation (ODE) \cite{liu2022flow}, it enables high-quality image generation under a few denoising steps, thus significantly improving both the efficiency and the realism of closed-loop simulation in real-world scenarios.

\begin{figure*}[t]
\vspace{2mm}
    \centering
    \includegraphics[width=0.99\textwidth]{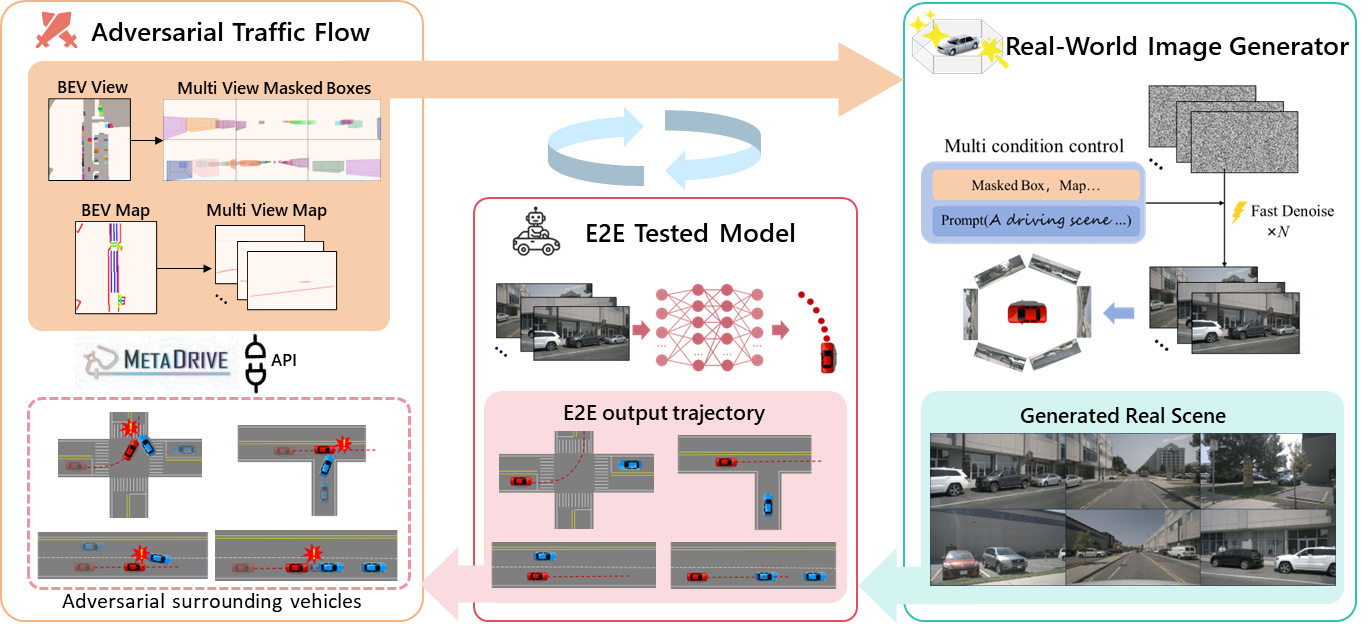}
    \caption{\textbf{Overview of the real-world adversarial closed-loop evaluation platform.} The platform integrates three key modules: Adversarial Traffic Flow, Real-World Image Generator, and E2E Tested Model. The Adversarial Traffic Flow controls surrounding vehicles that interact adversarially with the ego, providing traffic information to the Real-World Image Generator. The generator efficiently generates real-world images based on traffic information through flow matching. The generated images are passed as input to the E2E Tested Model, and the model's output is fed back to the Adversarial Traffic Flow, completing the closed-loop simulation.}
    \label{overrall}
\end{figure*}

This work proposes a real-world adversarial closed-loop evaluation platform for end-to-end autonomous driving, in which the real-world image generator collaborates with adversarial traffic flow to evaluate end-to-end models in corner cases.
Specifically, the generator correspondingly generates real-world images according to traffic conditions, such as lane markings and the distribution of surrounding objects. The flow matching-based model in the generator ensures efficiency and realism. 
Moreover, by filtering multimodal trajectories to select adversarial yet physically plausible ones, surrounding vehicles can efficiently construct safety-critical actions.
The experiment shows that generated images with low denoising steps in this platform outperform the baseline in terms of realism metrics. Our platform supports evaluating various end-to-end model instances. The results show that the adversarial traffic flow results in a sharp reduction of their scores and completion rates, effectively uncovering real-world corner cases that are difficult for the tested models to handle. 
Based on this design, the main contributions of this work are summarized as follows:

\begin{enumerate}
\item{We introduce the first closed-loop evaluating platform with real-world adversarial scenarios, 
which can evaluate the performance of various end-to-end models under adversarial interactions.}
% and provide specific evaluation metrics such as scores and completion rates

\item{The real-world image generator based on the traffic condition accordingly generates the real-world scene, in which the flow matching-based model using diffusion model priors can create high-quality real-world images with only a few denoising steps.}

\item{The adversarial traffic flow can efficiently identify and construct safety-critical interactions based on the adversarial scores of surrounding vehicles' multimodal trajectories. This trajectory selection mechanism is applicable to various tested models and traffic scenarios.}

% \item{The adversarial traffic flow can quickly create adversarial scenarios according to the adversarial scores of surrounding vehicles' multimodal trajectories. This adversarial policy is applicable to various tested models and traffic scenarios.}

% \item{ Evaluating various end-to-end models on this platform, . The platform creates for evaluating end-to-end models, promoting research on end-to-end model safety and robustness.}
\end{enumerate}

\section{METHODOLOGY}

% This section presents the proposed closed-loop platform for real-world adversarial scenarios. We first provide an overview of the platform, describing its operation and the information flow. We then introduce the implementation of adversarial traffic flow and the policy of adversarial surrounding vehicles, followed by the flow matching–based generator and the specific methods adopted in this work. Finally, we present the evaluation metrics.

This section presents the proposed closed-loop platform with real-world adversarial scenarios. We first provide an overview of the platform, describing its operation and the information flow. Then, we introduce the adversarial trajectories selection mechanism of surrounding vehicles, followed by the flow matching–based generator. Finally, we present the evaluation metrics.

\subsection{Overall Architecture}

As illustrated in Fig. \ref{overrall}, the proposed platform establishes a closed loop among three modules: Adversarial Traffic Flow, Real-World Image Generator, and the E2E Tested Model. The Adversarial Traffic Flow is driven by a simulator to provide a realistic vehicle physical simulation. In this work, we adopt the lightweight MetaDrive \cite{li2022metadrive} as the basis for Adversarial Traffic Flow, where obtain traffic conditions such as maps and surrounding vehicle distributions. This information is then passed to the Real-World Image Generator. The flow matching-based generator generates real-world scene images conditioned on the ego’s current traffic information, while ensuring compatibility with the input format of the end-to-end tested model. The E2E Tested Model takes the generated images as input and outputs the planned trajectory of the ego, which is subsequently fed back to the Adversarial Traffic Flow. The traffic flow then executes the ego's trajectory, thereby completing the closed-loop evaluation of real-world adversarial scenarios.

% To enhance the efficiency of exploring adversarial scenarios while avoiding unrealistic continuous confrontations, our platform adopts an epoch-based design. The Adversarial Traffic Flow loads real traffic segments and applies one segment per epoch, with each segment lasting for 9 seconds. Meanwhile, the conventional simulator driving the traffic flow can obtain signals such as drivable areas and vehicle collisions, making it easier to evaluate the tested end-to-end model. The flow matching–based Real-world Image Generator ensures high-quality images with a few denoising steps conditioned on the ego’s current traffic environment. Furthermore, by modifying the prompt descriptions of time and weather, it can flexibly produce real-world images in different styles.

\subsection{Adversarial Traffic Flow}

% Adversarial Traffic Flow is responsible for traffic and evaluation process management, as well as capturing platform operation logs. This section mainly introduces the structure of a single epoch and adversarial traffic policy.

Adversarial Traffic Flow is responsible for adversarial interaction and evaluation process management. This section mainly introduces the structure and adversarial trajectory selection mechanism.

\begin{figure}[t]
\vspace{2mm}
\centerline{\includegraphics[width=0.45\textwidth]{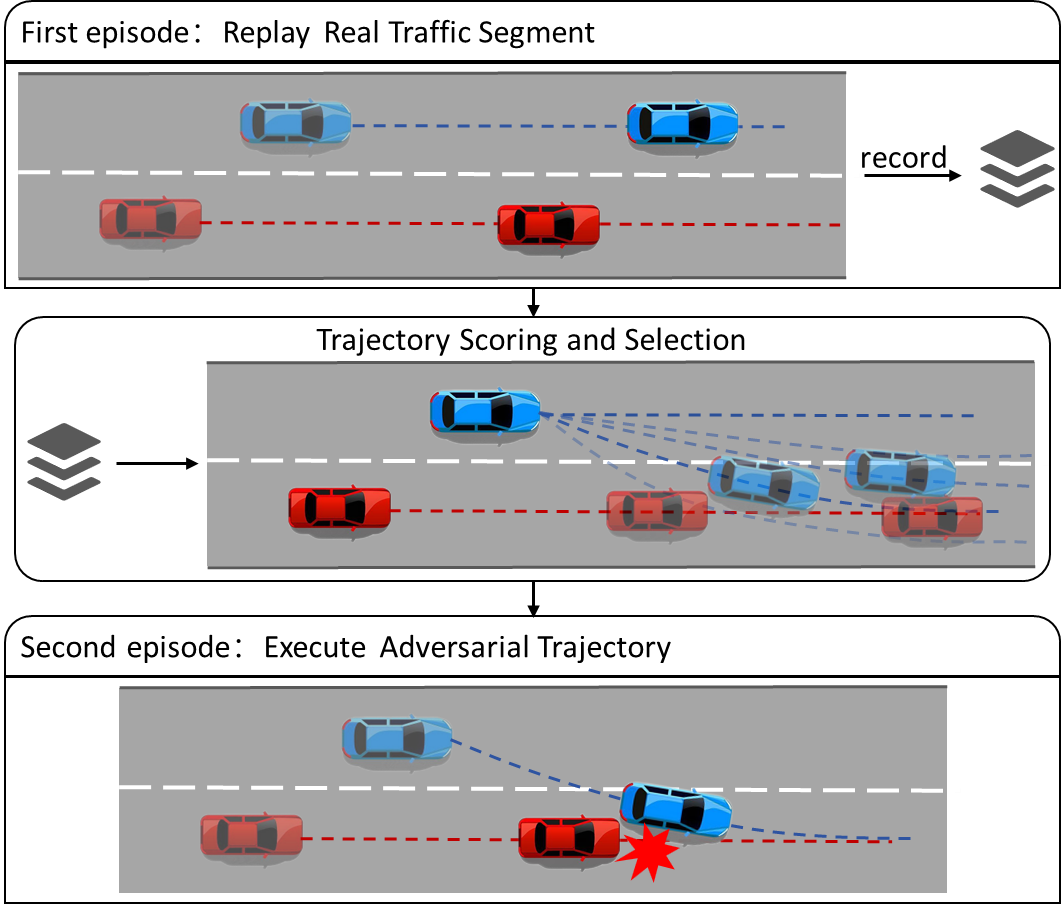}}
\caption{\textbf{Adversarial Trajectory of Surrounding Vehicle Selection Mechanism.} The method consists of two episodes. The first episode replays a steady traffic flow, and the trajectory of the tested model is recorded. Based on the recorded data, an adversarial and physically plausible trajectory of the surrounding vehicle is selected, and then this trajectory is applied in the second episode.}
\label{collision}
\end{figure}

\subsubsection{Two Episodes Architecture}

% Due to different end-to-end models having various preferences in metrics of speed and acceleration, etc, we adopt a two-episode adversarial scheme to ensure applicability across tested models, as illustrated in Fig. \ref{collision}. Specifically, our experiments are conducted on segments from real-world datasets. In the first episode, all vehicles except the ego replay log data, while the ego is controlled by the tested end-to-end model. During this process, adversarial vehicles record the complete trajectory of the ego and assess the tested model’s capability under regular traffic flow. Before the second episode, the adversarial vehicle queries a multimodal trajectory prediction model and gets multiple candidate trajectories. Based on the ego trajectory recorded in the first episode, it selects the trajectory with the highest adversarial score and executes it in the second episode. Two episodes form one epoch, executing one segment, i.e., one scenario. The loop shown in Fig. \ref{overrall} is executed across both episodes.

Due to different end-to-end models having various preferences, we adopt a two-episode architecture to ensure applicability across tested models, as illustrated in Fig. \ref{collision}. Specifically, our platform adopts an epoch-based design, and one epoch consists of two episodes, where both episodes load the same real-world traffic segment. The closed-loop shown in Fig. \ref{overrall} is executed across both episodes. In the first episode, all surrounding vehicles replay the segment, while the ego is controlled by the tested end-to-end model, and the complete trajectory of the ego is recorded. Before the second episode, the adversarial vehicle queries a multimodal trajectory prediction model trained on large-scale real-world traffic scenarios and gets multiple candidate trajectories. Based on the ego trajectory recorded in the first episode, it selects the trajectory with the highest adversarial score and executes it in the second episode. In this way, the trajectory of the adversarial surrounding vehicle can be adjusted according to the trajectory preferences of different tested models

\subsubsection{Adversarial Score}

To select surrounding vehicle trajectories that are both adversarial and physically plausible, we propose a multiplicative scoring function that jointly accounts for three key factors: (i) plausibility under the predictive model, (ii) likelihood of collision with the ego trajectory, and (iii) smoothness of motion. For each candidate trajectory $\tau_i$, the scoring function is defined as:

\begin{equation}
\begin{aligned}
\text{Score}(\tau_i) &= p_i \cdot (c_i)^{w_c} \cdot e^{-w_j J(\tau_i)} \\
c_i &= \gamma^{t_c(\tau_i)-1}
\end{aligned}
\label{score}
\end{equation}

Where $p_i$ denotes the prior probability output by the multimodal trajectory prediction model, ensuring adversarial trajectories remain consistent with realistic driving distributions; $c_i \in [0,1]$ represents the collision possibility with the ego, designed such that earlier collisions yield higher values of $c_i$; $t_c$ is the first collision time step of $\tau_i$, and $\gamma \in (0,1)$ is the decay factor, together ensuring that trajectories leading to earlier collisions are prioritized during selection; $J(\tau_i)$ denotes the normalized jerk penalty, which discourages abrupt acceleration changes and enforces physical plausibility; and $w_c$ and $w_j$ are weighting parameters balancing the relative importance of adversarial intensity and motion smoothness. The final adversarial trajectory is defined as:

% The trajectory with the highest score is selected as the adversarial trajectory. 
% With this design, the selected trajectory not only interacts with the ego adversarially but also avoids unrealistic or unstable motion patterns.

% The final adversarial trajectory is defined as:

\begin{equation}
\tau^* = \arg\max_i \; \text{Score}(\tau_i)
\label{score_best}
\end{equation}

% With this design, the selected trajectory not only remains consistent with the prediction model but also poses adversarial challenges to the ego, while avoiding unrealistic or unstable motion patterns.

With this design, the selected trajectory not only interacts with the ego adversarially but also avoids unrealistic or unstable motion patterns.

\subsection{Real-World Image Generator}

% Diffusion models have demonstrated outstanding performance in high-quality generation tasks, particularly the foundation model of Stable Diffusion\cite{rombach2022high}, which exhibits remarkable adaptability to downstream tasks. Although methods such as DDIM\cite{song2020denoising} attempt to reduce the number of denoising steps, fewer denoising steps are often accompanied by a decline in generation quality. To address this, we adopt the ODE-based flow matching approach, employing the more computationally efficient Euler method for fast denoising, as shown in Eq. \ref{flow_denoise}. This enables us to preserve image quality even with a reduced number of denoising steps.

% Diffusion models have demonstrated outstanding performance in generation tasks, particularly the foundation model of Stable Diffusion\cite{rombach2022high}. Although methods such as DDIM\cite{song2020denoising} attempt to reduce denoising steps, fewer denoising steps are often accompanied by lower quality. To address this, we adopt the ODE-based flow matching approach, employing the more computationally efficient Euler method for fast denoising. This enables us to preserve image quality even with a few denoising steps. 第一次修改

Diffusion models have demonstrated outstanding performance in generation tasks, particularly the foundation model of Stable Diffusion\cite{rombach2022high}. Although methods such as DDIM\cite{song2020denoising} attempt to reduce denoising steps, fewer denoising steps are often accompanied by lower quality. To address this, we adopt the ODE-based flow matching approach, employing the more computationally efficient Euler method for fast denoising and high image quality.

% \begin{equation}
% x_{t+\Delta t} = x_t +\Delta t\cdot v^{F}_{\theta}(x_t, t), \forall t \in [0, 1)
% \label{flow_denoise}
% \end{equation}

% \begin{equation}
% x_{t+\Delta t} = x_t +\Delta t\cdot v
% \label{flow_denoise}
% \end{equation}

\subsubsection{Flow Matching}

% Existing flow-matching foundation models often require a large number of parameters, which makes them highly resource-intensive. In contrast, the efficient architecture and rich ecosystem of Stable Diffusion provide promising insights for flow matching. Inspired by \cite{schusterbauer2025diff2flow}, we adopt a linear interpolation approach to seamlessly transfer the priors of pretrained diffusion models into velocity field prediction within flow-matching models, as shown in Fig. \ref{flow}. In this way, we leverage the strong priors of diffusion models while simultaneously benefiting from the efficient inference advantage of flow matching.

% Existing flow-matching foundation models often require a large number of parameters. In contrast, the efficient architecture and rich ecosystem of Stable Diffusion provide promising insights for flow matching. Inspired by \cite{schusterbauer2025diff2flow}, we adopt a linear interpolation approach to transfer the priors of pretrained diffusion models into flow-matching models, as shown at the top of Fig. \ref{flow}. In this way, we leverage the priors of diffusion models while benefiting from the efficient inference advantage of flow matching. 第一次修改

\begin{figure*}[t]
    \centering
    \includegraphics[width=0.85\textwidth]{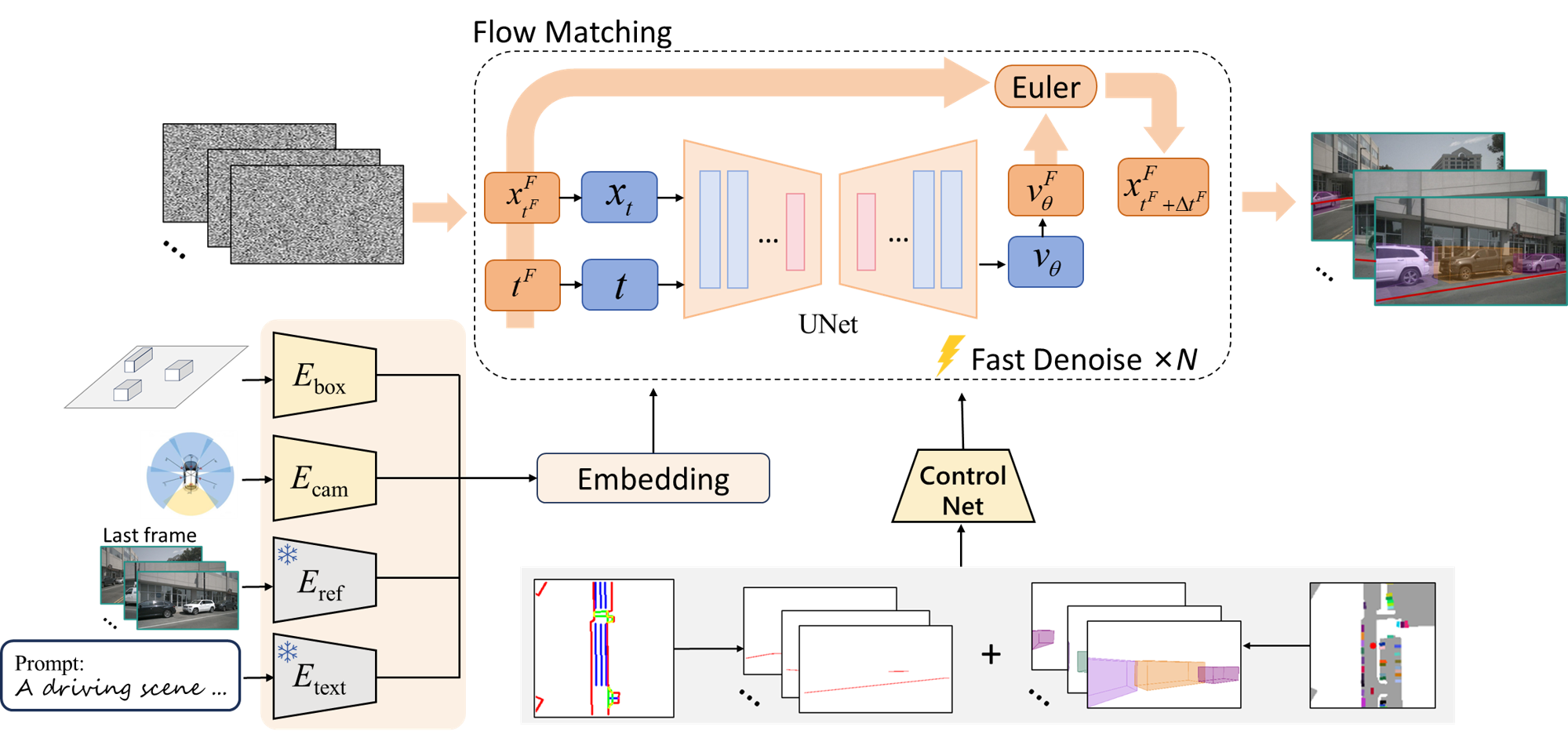}
    \caption{\textbf{Overview of the Real-World Image Generator.} The backbone network of flow matching is a UNet, which leverages diffusion priors through linear transformation. Information projected into the camera view is injected via ControlNet, while other conditional information is incorporated through attention mechanisms.}
    \label{flow}
\end{figure*}

Existing flow-matching models often require large parameter counts. In contrast, Stable Diffusion offers an efficient architecture and a rich ecosystem. Inspired by \cite{schusterbauer2025diff2flow}, we adopt a linear interpolation approach to transfer diffusion priors into flow-matching models, as shown at the top of Fig. \ref{flow}, thereby combining pretrained priors with the efficient inference of flow matching.

The outputs of diffusion models generally take three common forms: noise, $x_0$, and $v_\theta$, which are equivalent. In our analysis, we adopt the $v_\theta$. The noise-adding equation of the diffusion model and the parameterization of $v_\theta$ are defined in Eq. \ref{dm}, $t$ is the diffusion timestep.

\begin{equation}
\begin{bmatrix}
x_t \\
v_\theta
\end{bmatrix}
=
\begin{bmatrix}
\alpha_t & \sigma_t \\
-\sigma_t & \alpha_t
\end{bmatrix}
\begin{bmatrix}
x_0 \\
x_T
\end{bmatrix}.
\label{dm}
\end{equation}

% In the diffusion scheduler, $\alpha_t$ and $\sigma_t$ satisfy $\alpha_t^2 + \sigma_t^2 = 1$, which allows us to inversely derive $\hat{x_0}$ and $\hat{x_T}$. The noise injection and velocity $v^{F}_{\theta}$ definition and conversion relationship in flow matching are given in Eq. \ref{fm}, $t^F$ is the flow matching timestep. It should be noted that, following the conventions in diffusion and flow matching, $x_0$ and $x^{F}_1$ denote the original data samples, whereas $x_T$ and $x^{F}_0$ represent random Gaussian noise.

In diffusion model, $\alpha_t$ and $\sigma_t$ satisfy $\alpha_t^2 + \sigma_t^2 = 1$, which allows us to inversely derive $x_0$ and $x_T$. The noise-adding equation and velocity $v^{F}_{\theta}$ definition in flow matching and conversion relationship are given in Eq. \ref{fm}, $t^F$ is the flow matching timestep. It should be noted that, following the conventions in diffusion and flow matching, $x_0$ and $x^{F}_1$ denote the original data samples, whereas $x_T$ and $x^{F}_0$ represent random Gaussian noise.

\renewcommand{\arraystretch}{1.5} % 调整行距倍数
\begin{align}
& \left\{ \begin{array}{l}
x_{t^{F}}^{F} = t^{F}x^{F}_1 + (1-t^{F}) x^{F}_0 \\
v^{F}_{\theta} =x^{F}_1-x^{F}_0 = x_0 - x_T  \\ 
\;\;\;\;\; = (\alpha_{t}-\sigma_{t})x_t - (\alpha_{t}+\sigma_{t})v_{\theta}
\end{array} \right.
\label{fm}
\end{align}

This conversion originates from linear algebra matrix inversion, which avoids retraining a nonlinear mapping from $v_{\theta}$ to the flow matching velocity $v^{F}_\theta$ through the network. This significantly reduces the training complexity while fully leveraging the prior knowledge embedded in the pretrained model. Moreover, $t^F$ is defined as a continuous time in $[0,1]$, whereas diffusion $t$ usually adopt discrete steps (e.g., $0$–$999$). The model still receives $(t, x_t)$, thus requiring a linear interpolation scheme to map between $x^{F}_{t^F}, t^{F}$ and $t, x_t$. Based on the noise-adding equation in Eq. \ref{dm} and Eq. \ref{fm}, the conversion can be constructed shown in Eq. \ref{dm2fm}, and Eq. \ref{flow_denoise} demonstrates the Euler-based denoising. For the loss function, we use the mean squared error (MSE) between the $v^F_\theta $and the $x^F_1 - x^F_0$, as shown in Eq. \ref {loss}.

% \renewcommand{\arraystretch}{1.5} % 调整行距倍数
% \begin{align}
% & \left\{ \begin{array}{l}
% \hat{t^F} = \alpha_t /(\alpha_t+\sigma_t) \in [0, 1) \\
% \hat{x_{t^{F}}^{F}} = \hat{t^F}\cdot x_1 + (1-\hat{t^F})\cdot x_0 = x_t/(\alpha_t+\sigma_t)
% \end{array} \right.
% \label{dm2fm}
% \end{align}

\renewcommand{\arraystretch}{1.5} % 调整行距倍数
\begin{align}
& \left\{ \begin{array}{l}
{t^F} = \alpha_t /(\alpha_t+\sigma_t) \in [0, 1) \\
{x_{t^{F}}^{F}} = {t^F}\cdot x^F_1 + (1-{t^F})\cdot x^F_0 = x_t/(\alpha_t+\sigma_t)
\end{array} \right.
\label{dm2fm}
\end{align}
\begin{equation}
x^{F}_{t^{F}+\Delta t^{F}} = x^{F}_{t^{F}} +\Delta t^{F}\cdot v^{F}_{\theta}
\label{flow_denoise}
\end{equation}
\begin{equation}
\mathcal{L} = MSE((x^F_1-x^F_0), v^{F}_\theta)
% \mathcal{L} = \big\| (x^{F}_{1} - x^{F}_{0}) - v^{F}_{\theta} \big\|_2^2
\label{loss}
\end{equation}

\subsubsection{Condition Control}
% To enable controllability over critical traffic elements in the generated images, we project the 3D bounding boxes of surrounding objects onto each camera view. To ensure consistency across viewpoints, each object is assigned a unique ID. When projecting the same object onto different views, the corresponding projection regions are masked according to this ID, as illustrated by different colors in Fig. \ref{flow}. A similar approach is applied to project lane markings from the map into the various camera views. To maintain temporal consistency, we follow the approach in \cite{yang2024drivearena} by using the last frame as a reference, which provides better control over weather and surrounding street scenes. Additional auxiliary conditions include low-dimensional information such as camera intrinsic and extrinsic parameters $(K, R, T)$ and the coordinates of the 3D bounding boxes of surrounding objects.

To enable controllability over critical traffic elements in the generated images, we project the 3D bounding boxes of surrounding objects onto each camera view. To ensure consistency across viewpoints, each object is assigned a unique ID. When projecting the same object onto different views, the corresponding projection regions are masked according to this ID. A similar approach is applied to project lane markings into the various camera views. To maintain temporal consistency, we follow the approach in \cite{yang2024drivearena} by using the last frame as a reference, which provides better control over weather and street scenes. Additional auxiliary conditions include low-dimensional information such as camera intrinsic and extrinsic parameters and the coordinates of the 3D bounding boxes of surrounding objects.

% We primarily incorporate conditional information through ControlNet and attention mechanisms, as illustrated in Fig. \ref{flow}. Both the reference image and prompts are encoded using a frozen CLIP \cite{radford2021learning} model. Low-dimensional information is processed via a trainable MLP, and these features are injected into the UNet through cross-attention. Map and surrounding objects projected into the camera views are injected into the UNet using a trainable ControlNet. We adopt the features encoded by a frozen VAE \cite{kingma2013auto} as the denoising backbone, which significantly reduces computational costs.

% We primarily incorporate conditional information through ControlNet and attention mechanisms, as illustrated in Fig. \ref{flow}. The features of the reference image, prompts, and low-dimensional information are injected into the UNet through cross-attention. By modifying the prompt descriptions of time and weather, it can flexibly produce real-world images in different styles. Map and surrounding objects projected into the camera views are injected into the UNet using a trainable ControlNet. 第一次修改

We incorporate conditions through ControlNet and attention mechanisms, as illustrated in Fig. \ref{flow}. The features of the reference image, prompts, and low-dimensional information are injected into the UNet through cross-attention. Based on the prompts of various times and weather, the generator can produce real-world images in different styles. Map and surrounding objects projected into the camera views are injected into the UNet using a trainable ControlNet.

\subsection{Evaluation Metrics}

% \subsubsection{Vehicle Control}

% The Adversarial Traffic Flow is built upon the MetaDrive simulator, which provides the api for both longitudinal and lateral control. Since most end-to-end autonomous driving models trained on real-world data primarily output trajectories, we employ a PID controller within the simulator to track these trajectories. Apart from the adversarial vehicles and the ego, all background vehicles are controlled directly by MetaDrive.

% \subsubsection{Evaluate Metrics}

% Our platform will quantitatively score the performance of the tested model under adversarial traffic flow. In the evaluation of autonomous driving, common single-frame metrics include no collisions (NC), drivable area compliance (DAC), ego progress (EP), time-to-collision (TTC), and comfort (C). For the EP, since we adopt real-world traffic segments, the ground-truth ego trajectory can be used as a reference, and the route completion (RC) of each segment can be derived. We also record the termination reason of each episode and determine whether the ego collides with surrounding vehicles, so the segment completion (SC) can be calculated after running all segments.  第一次修改

Our platform will quantitatively score the performance of the tested model under adversarial traffic flow. In the evaluation of autonomous driving, common single-frame metrics include no collisions (NC), drivable area compliance (DAC), ego progress (EP), time-to-collision (TTC), and comfort (C). Since we adopt real-world traffic segments, the ground-truth ego trajectory can be used as a reference, and the route completion (RC) can be derived. We also record the termination reason of each episode and determine whether the ego collides with surrounding vehicles, so the segment completion (SC) can be calculated after running all segments.

\begin{align}
\text{PDMS} &=
\underbrace{\left( \prod_{m \in \{\text{NC}, \text{DAC}\}} \text{score}_m \right)}_{\text{penalties}}
\times \notag \\
&\quad
\underbrace{\left(
\frac{
\sum\limits_{w \in \{\text{EP}, \text{TTC}, \text{C}\}}
  \text{weight}_w \times \text{score}_w}
{\sum\limits_{w \in \{\text{EP}, \text{TTC}, \text{C}\}}
  \text{weight}_w}
\right)}_{\text{weighted average}}
\label{pdms}
\end{align}

% The single-frame metrics indicate discretely different aspects of driving performance. NAVSIM\cite{dauner2024navsim} proposed the PDMS metric by integrating these metrics, which has been widely adopted in both academia and industry, as shown in Eq. \ref{pdms}. Following this definition, we compute the single-frame PDMS and then average it across the entire segment. The final closed-loop driving score (DS) is obtained by multiplying the average PDMS by the RC.

The single-frame metrics indicate different aspects of driving performance. NAVSIM\cite{dauner2024navsim} proposed the PDMS by integrating these metrics, which has been widely adopted in both academia and industry, as shown in Eq. \ref{pdms}. Following this definition, we compute the average PDMS across the entire segment. The final closed-loop driving score (DS) is obtained by multiplying the average PDMS by the RC.

\section{EXPERIMENTS}

This section introduces the setups of the experiments, the realism of the images generated by Real-World Image Generator, and the experimental results of different tested end-to-end models on our platform.

\subsection{Experimental Setups}

\begin{table*}[t]
\vspace{2mm}
\centering
\resizebox{\textwidth}{!}{
\begin{tabular}{l|cccc||cccc||cccc}
\hline
& \multicolumn{4}{c||}{FID↓ / FVD↓} & \multicolumn{4}{c||}{LPIPS↓} & \multicolumn{4}{c}{SPI↓} \\
& {20 steps} & {10 steps} & {5 steps} & {3 steps} & {20 steps} & {10 steps} & {5 steps} & {3 steps} & {20 steps} & {10 steps} & {5 steps} & {3 steps} \\ \hline
{MagicDrive\cite{gao2023magicdrive}}& {13.37 / 2.43}& {14.32 / 2.66} & {18.49 / 3.38} & {36.59 / 5.37} & {0.433} & {0.432} & {0.439} & {0.460} & {2.38} & {1.48} & {1.02} & {0.83} \\
{DriveArena\cite{yang2024drivearena}}& {14.17 / 2.58} & {15.65 / 2.70} & {20.00 / 3.33}    & {32.99 / 5.08} & {0.354} & {0.353} & {0.359} & {0.382}  & {2.44} & {1.53} & {1.06} & {0.85} \\
{\textbf{Ours}}& {\textbf{12.05 / 2.24}} & {\textbf{12.92 / 2.34}} & {\textbf{15.64 / 2.73}} & {\textbf{23.63 / 3.76}} & {\textbf{0.323}} & {\textbf{0.319}} & {\textbf{0.320}} & {\textbf{0.338}} & {2.43} & {1.52} & {1.04} & {0.85} \\ \hline

\end{tabular}}
\caption{\textbf{Comparison of the FID, FVD, LPIPS, and SPI.} The Real-World Image Generator demonstrates an advantage in image quality under a few denoising steps, without increasing computational cost.}
\label{fid}
\end{table*}

\subsubsection{Real-World Image Generator}
We train and validate the image generator on the nuScenes dataset 
(1,000 driving scenes), using 850 for training and validation and 150 for testing,
and adopt Stable Diffusion 1.5 with the UNet architecture. For condition control, we freeze the pretrained CLIP \cite{radford2021learning} for prompt and reference image encoders while training the remaining components. Our baselines are DDIM-based image generation models, MagicDrive \cite{gao2023magicdrive} and DriveArena \cite{yang2024drivearena}, both built upon Stable Diffusion 1.5, which are representative controllable driving image generation methods.

% The generator is trained and evaluated using a common resolution of $224 \times 400$. To match the original resolution of $900 \times 1600$ in the nuScenes dataset, we employ the Camixersr\cite{wang2024camixersr} to upsample the images before feeding them into the end-to-end model. Fast diffusion models based on DDIM typically adopt 50-–100 denoising steps, while the flow matching method can achieve high image quality with a few denoising steps. We employ a fast denoising scheme with 10 steps and set the condition guidance scale to 2.0. For training, we use 8 NVIDIA A800 GPUs with an initial learning rate of $8 \times 10^{-5}$, which is gradually decayed during training. The batch size is set to 4, and the model is trained for 40 epochs.

Fast diffusion models based on DDIM typically adopt 50-100 denoising steps, while our image generator can achieve high image quality with a few denoising steps. We employ a fast denoising scheme with 10 steps and set the condition guidance scale to 2.0. For training, we use 8 NVIDIA A800 GPUs with an initial learning rate of $1 \times 10^{-4}$, which is gradually decayed during training. The batch size is set to 4, and the model is trained for 20 epochs.

\subsubsection{Closed-loop Setups}

We employ the Waymo Open Motion Dataset(WOMD) \cite{ettinger2021large} and convert it into the format of MetaDrive for replay, where a diverse collection of real traffic scene segments is gathered. The adversarial vehicle adopts the DenseTNT \cite{gu2021densetnt} multimodal trajectory prediction model, which outputs 32 candidate trajectories with prior probabilities. For the tested end-to-end models, we choose UniAD and VAD, both representative approaches in end-to-end autonomous driving. The closed-loop runs at 2Hz, while Adversarial Traffic Flow runs at 10Hz.

\begin{figure}[t]
\centerline{\includegraphics[width=0.48\textwidth]{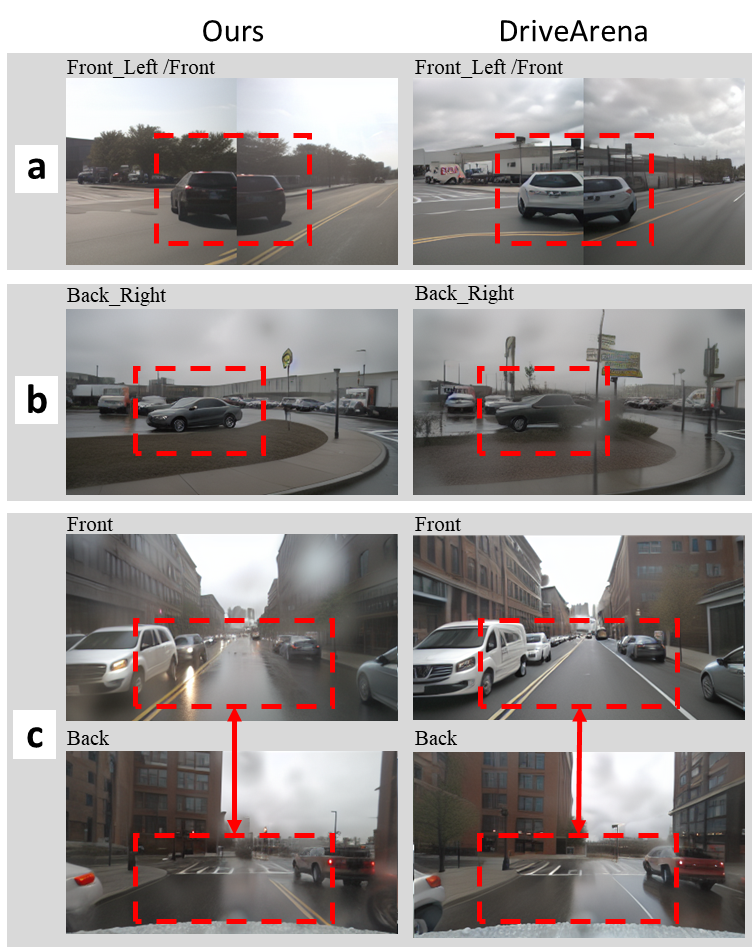}}
\caption{\textbf{Comparison of generated image quality.} We provide three sets of example images: a, b, and c. In sub-figures a and b, the elements in the red boxes clearly show that our generator generates higher-quality images. In sub-figure c, it can be observed that both the front and back views are consistently rainy, while the baseline shows noticeable differences between the two views.}
\label{qua}
\end{figure}

\subsection{Image Generation Validation}

% To quantitatively evaluate the visual quality of the generated images, we adopt two common metrics: Fréchet Inception Distance (FID) and Learned Perceptual Image Patch Similarity (LPIPS). Specifically, FID measures the distributional discrepancy between generated and real images in the feature space, where lower values indicate higher fidelity; LPIPS evaluates perceptual similarity by comparing deep feature representations, thereby providing an assessment consistent with human perception. We conduct comparative experiments using the released weights of MagicDrive and DriveArena. To further validate the superiority of the flow matching under low denoising steps, we additionally report results with $5$ steps and $3$ steps denoising. Moreover, we record the seconds per item (SPI), i.e., the time required to complete a full denoising process under the same computational resources. As summarized in TABLE \ref{fid}, our results demonstrate that under low denoising steps, both FID and LPIPS outperform the baselines without incurring additional runtime, confirming that the improvements are not achieved by scaling up the model or consuming extra resources.

To quantitatively evaluate the visual quality of the generated images, we adopt the common metrics: Fréchet Inception Distance (FID), Fréchet Video Distance (FVD) and Learned Perceptual Image Patch Similarity (LPIPS). Specifically, FID and FVD measure the distributional discrepancy between generated and real data in feature space—FID for single images and FVD for videos by measuring temporal consistency across frames; LPIPS evaluates perceptual similarity by comparing deep feature representations, thereby providing an assessment consistent with human perception. We conduct comparative experiments using the official pretrained weights of MagicDrive and DriveArena, and perform evaluation on the nuScenes validation split. To further validate the superiority of our generator under low denoising steps, we additionally report results with both low-step (5 and 3) and high-step (20) denoising settings. Moreover, the seconds per item (SPI) is recorded, i.e., the time required for a complete denoising process under the same computational resources. We use validation set labels from nuScenes to generate approximately 36,000 images for experimentation. 
As summarized in TABLE \ref{fid}, our results demonstrate that under low denoising steps, FID, FVD and LPIPS outperform the baselines without incurring additional runtime, confirming that the improvements are not achieved by scaling up the model or consuming extra resources. Moreover, with only 10 denoising steps, our image generator already outperforms the baselines using 20 steps, indicating that we can reduce computational cost and achieve the desired quality under lower time consumption than the baseline. Considering efficiency and quality, we select 10 steps for subsequent experiments.

% To validate the effectiveness of key elements such as vehicles and lane markings in the generated images, we use the validation set labels from nuScenes to generate images. 
To validate the effectiveness of key elements such as vehicles and lane markings in the generated images, UniAD works as the evaluator and is performed on these generated images and computes the perception-related metrics, including 3d object detection(3DOD), bird’s-eye view(BEV) segmentation. 
The experimental results are shown in TABLE \ref{uniad}. 
% The results indicate that the gap between the generated images by our model and the original images is minimal in most perception metrics. 
The results indicate that the gap between our generated images and the original images is minimal in most perception metrics. 
% The results show that our generated images closely match the original images across most perception metrics.
This also shows that our generator is highly controllable and can generate images strictly according to conditions. 
Meanwhile, we present some generated results, as shown in Fig. \ref{qua}. 
From the three sets of generated images, it can be seen that our generation quality is obviously higher. 
% In sub-figure c of Fig. \ref{qua}, our generator can generate road conditions with rainy weather for both the front and back views, while the baseline exhibits noticeable differences between the front and back views.

\subsection{Adversarial Closed-loop Experiments}

% We conducted tests on two representative end-to-end models, UniAD and VAD, using the official code and pre-trained weights. We experimented with various adversarial scenarios, including cut-ins, sudden braking of leading vehicles, intersection negotiations, and rapid merges etc. The closed-loop ablation results with and without adversarial traffic are summarized in TABLE \ref{test}. Initially, in a traffic environment without adversarial surrounding vehicles, both UniAD and VAD achieved both SC and RC above 0.8, with good performance across all metrics, demonstrating that the models can operate well in a steady closed-loop environment. However, after introducing adversarial surrounding vehicles, the comprehensive scores PDMS and DS dropped significantly by 30 \%--50 \%, and the completion rates SC and RC also decrease significantly. These results indicate that adversarial surrounding vehicles create numerous corner cases for the end-to-end models. This notable performance decline highlights the weakness of current end-to-end driving models when facing adversarial traffic participants. Although UniAD and VAD can perform well under steady driving conditions, they still lack the capability to handle rare and safety-critical interactions. This also exposes the limitations of purely imitation learning in out-of-distribution scenarios.第一次修改

We conduct tests on two representative end-to-end models, UniAD and VAD, using the official code and pre-trained weights. 
% We experiment over 500 real driving segments from WOMD with various adversarial scenarios, including cut-ins, sudden braking of leading vehicles, intersection games, and rapid merges etc. 
We experiment over 500 real driving segments from WOMD, in which diverse adversarial interactions arise through multimodal trajectories filtering, including cut-ins, sudden braking of leading vehicles, intersection conflicts, and rapid merges, etc.
The closed-loop ablation results with and without adversarial traffic are summarized in TABLE \ref{test}. Initially, in a traffic environment without adversarial surrounding vehicles, both UniAD and VAD achieved both SC and RC above 0.8, with good performance across all metrics. However, after introducing adversarial surrounding vehicles, the comprehensive scores PDMS and DS dropped significantly by 30 \%--50 \%, and the completion rates SC and RC also decreased significantly. These results indicate that adversarial surrounding vehicles create numerous corner cases to challenge the end-to-end models. This notable performance decline highlights the weakness of current end-to-end driving models when facing adversarial traffic participants. Although UniAD and VAD can perform well under steady driving conditions, they still lack the capability to handle safety-critical interactions. 
% This also exposes the limitations of imitation learning in out-of-distribution scenarios.

\begin{table}[t]
\vspace{2mm}
\centering
\resizebox{\columnwidth}{!}{
\begin{tabular}{l|cc|cccc}
\hline
& \multicolumn{2}{c|}{3DOD↑} & \multicolumn{4}{c}{BEV Segmentation mIoU (\%)↑}   \\ \cline{2-7} 
\multirow{-2}{*}{Data Source }& {mAP}   & {NDS}   & {Lanes} & {Drivable} & {Divider} & {Crossing} \\ \hline
\rowcolor[HTML]{E0E0E0} 
{Original}&{37.93}&{43.67}&{31.31}&{69.15}&{25.94}&{14.37}\\
{MagicDrive\cite{gao2023magicdrive}}&{14.54}&{27.00}&{23.53}&{54.63}&{18.93}&{6.48}\\
{DriveArena\cite{yang2024drivearena}}&{16.36}&{28.83}&{28.37}&{62.54}&{22.69}&{\textbf{10.84}}\\
{\textbf{Ours}}&{\textbf{18.46}}&{\textbf{29.93}}&{\textbf{28.52}}&{\textbf{63.07}}&{\textbf{23.05}}&{10.75}\\ \hline
\end{tabular}
}
\caption{\textbf{Comparison of fidelity in generating important traffic information.} Generated images are input into UniAD to compare perception results. Original represents using the original real image as input. The \textbf{bolded result} denotes the smallest gap between the perception results of the generated images and the original real images.}
\label{uniad}
\end{table}

\begin{table*}[t]
\vspace{2mm}
\centering
\begin{tabular}{l|c|ccccc
>{\columncolor[HTML]{E0E0E0}}c 
>{\columncolor[HTML]{E0E0E0}}c 
>{\columncolor[HTML]{E0E0E0}}c 
>{\columncolor[HTML]{E0E0E0}}c }
\hline
Tesed Model & { Traffic Condition} & {NC↑}&{DAC↑}&{TTC↑}&{C↑}&{EP↑}&{PDMS↑}&{RC↑}&{DS↑}&{SC↑}\\ \hline
&{w/o adv}&{0.892}&{0.976}&{0.851}&{0.798}&{0.752}&{0.721}&{0.905}&{0.683}&{0.867}\\
\multirow{-2}{*}{UniAD}&{w/ adv}&{0.550}&{0.978}&{0.661}&{0.747}&{0.742}&{0.430 \footnotesize\textsubscript{-0.291}}&{0.679 \footnotesize\textsubscript{-0.226}}&{0.356 \footnotesize\textsubscript{-0.327}}&{0.398 \footnotesize\textsubscript{-0.469}}\\ \hline
  &{w/o adv}&{0.854}&{0.929}&{0.835}&{0.999}&{0.737}&{0.674}&{0.848}&{0.615}&{0.806}\\
\multirow{-2}{*}{VAD}&{w/ adv}&{0.574}&{0.941}&{0.733}&{1.000}&{0.711}&{0.446 \footnotesize\textsubscript{-0.228}}&{0.648 \footnotesize\textsubscript{-0.200}}&{0.340 \footnotesize\textsubscript{-0.275}}&{0.408 \footnotesize\textsubscript{-0.398}}\\ \hline
\end{tabular}
\caption{\textbf{Ablation results of adversarial closed-loop evaluation.} The metrics with a \colorbox[HTML]{E0E0E0}{gray background} are comprehensive closed-loop scores and completion rates, and the numbers in small font represent the scores or completion rates reduction in adversarial traffic flow compared to steady traffic flow. The adv in the table represents adversarial traffic.}
\label{test}
\end{table*}

\begin{figure*}[t]
% \vspace{2mm}
    \centering
    \includegraphics[width=0.99\textwidth]{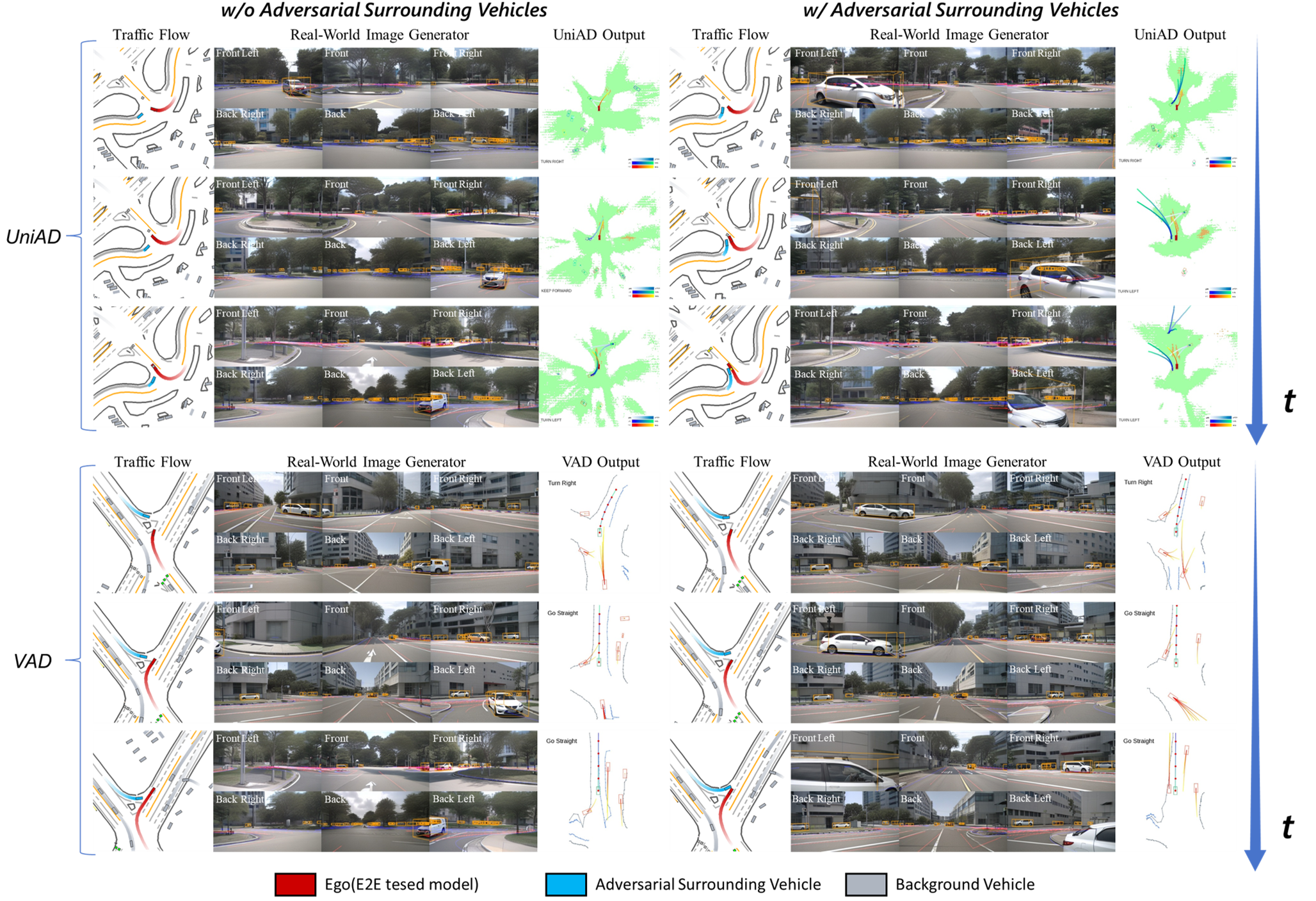}
    % \caption{\textbf{A typical case in adversarial closed-loop evaluating.} The top and bottom sections show the performance of UniAD and VAD, and we capture three key frames from the interaction. In each cell, the left side displays the ground truth traffic flow extracted from MetaDrive, with the red block representing the ego, the blue block representing the adversarial vehicle, and the other as background vehicles. The center shows the generated image from the Real-World Image Generator. The right side displays the output of the tested end-to-end model.}
    \caption{\textbf{A typical case in adversarial closed-loop evaluating.} The top and bottom sections show the performance of UniAD and VAD, and we capture three key frames from the interaction. In each cell, the left side displays the ground truth traffic flow extracted from MetaDrive. The center shows the generated image from the Real-World Image Generator. The right side displays the output of the tested end-to-end model.}
    \label{case_adv}
\end{figure*}

To further demonstrate the impact of adversarial vehicles, we present a typical case study. Fig. \ref{case_adv} shows the evaluation scene, including the traffic flow, the generated images, and the output of the tested model. From the results, it can be observed that without adversarial vehicles, both UniAD and VAD pass successfully. However, when adversarial surrounding vehicles are introduced, although UniAD’s planned trajectory exhibits an evasive tendency, it fails to avoid a collision. And VAD also fails to make an avoidance maneuver and causes a collision in the intersection game scenario. It can be observed that the generated images have high fidelity, almost indistinguishable from real traffic scenes. The adversarial vehicles maintain consistency in both temporal and perspectival aspects, and their direction and positioning are generated accurately. Therefore, it can be concluded that the failure of the ego is due to the safety-critical corner case created by the adversarial interaction of the surrounding vehicles rather than image quality. This example demonstrates that our adversarial platform can effectively expose the limitations and potential problems of end-to-end models while maintaining a high level of scenario realism.

\section{CONCLUSIONS}

% Adversarial scenarios in the real world are difficult to collect, and the datasets gathered are often unable to facilitate closed-loop testing. To address this, we propose an innovative real-world adversarial closed-loop evaluating platform that combines flow matching-based image generation with adversarial surrounding vehicles to perform stress testing on end-to-end autonomous driving systems. The Flow Matching method ensures the quality of generated images under low denoising steps, significantly improving the efficiency of real-world closed-loop simulation. Adversarial traffic flow is explored through a two-episode design to identify the most adversarial trajectories for different end-to-end models. Experimental results show that by leveraging fast, high-quality image generation and adversarial traffic scenarios, we can identify scenarios that existing models fail to cover. Although representative end-to-end models like UniAD and VAD perform well under regular driving conditions and open-loop testing, they exhibit vulnerabilities in adversarial closed-loop scenarios, with significant drops in various driving metrics. Therefore, the proposed platform provides a valuable tool for performance evaluating of end-to-end autonomous driving systems.

%Adversarial interaction is an effective method for generating corner cases.
This paper introduces an innovative real-world adversarial closed-loop evaluating platform that integrates a Real-World Image Generator and Adversarial Traffic Flow to perform adversarial interaction testing on E2E Tested Models trained on the real-world dataset. The flow matching method in the generator ensures the quality of generated real-world images under low denoising steps, significantly improving the efficiency of real-world closed-loop evaluation. Adversarial traffic flow is explored through a two-episode design to select the most adversarial surrounding vehicles' trajectories for different end-to-end tested models. Experimental results show that the realism metrics of the images generated by our platform are better than those of the baseline, and the generator has strong controllability over key traffic elements in the generated images. By the fast, high-quality image generator and adversarial traffic flow, we can construct safety-critical scenarios that tested models fail to handle. Although representative end-to-end models like UniAD and VAD perform well under steady traffic conditions, they exhibit potential issues in adversarial closed-loop scenarios, with significant drops in scores and completion rate. In conclusion, the proposed platform provides a valuable tool for evaluating end-to-end autonomous driving in corner cases.

\bibliographystyle{IEEEtran}
\bibliography{reference}

\end{document}